# Optimizing the Passenger Flow for Airport Security Check

Yuxin Wang[1], Fanfei Meng[1], Xiaotian Wang[2], Chaoyu Xie[2]

[1] Northwestern University, Evanston, IL, USA
[2] University of Science and Technology of China, Hefei, China

yuxin.wang@northwestern.edu

**Abstract**. Due to the necessary security for the airport and flight, passengers are required to have strict security check before getting aboard. However, there are frequent complaints of wasting huge amount of time while waiting for the security check. This paper presents a potential solution aimed at optimizing gate setup procedures specifically tailored for Chicago O'Hare International Airport. By referring to queueing theory and performing Monte Carlo simulations, we propose an approach to significantly diminish the average waiting time to a more manageable level. Additionally, our study meticulously examines and identifies the influential factors contributing to this optimization, providing a comprehensive understanding of their impact.

1. **Introduction**

*1.1. Background*

Transportation Security Administration (TSA) comes under sharp criticism for extremely long waiting lines, and several changes have been made to take care of the highly congested airports. As one of the biggest airports in the US and in the world, Chicago O'Hare International Airport (ORD) is suffering from the extremely long waiting lines for the security check. There are always a lot of complaints by passengers, and some of them even fail to get on the plane due to the long security check time. In addition to the issues at ORD, there are also incidents of unexplained and unpredicted long lines at other airports, as we can see on the news and the Internet. This high variance in checkpoint lines can be extremely costly to passengers as they decide between arriving unnecessarily early or potentially missing their scheduled flight.

We estimated the average passenger throughput per day for each terminal in ORD, and the throughput for busy days and the vacancy days. It shows that the total number of passengers in ORD in 2015 is 76,949,336, and 4 terminals are mainly used for civil transportation. If we assume that the airport works for 18 hours per day, then we get the average flow of passengers in the whole year: 2928 persons are coming for security check every hour in each terminal.

However, dealing with 2928 is not an easy task for a single terminal in one hour. After cautiously analyzing the data provided by ORD, we acquire the expectation of the time needed for a person to get through the Zone A, document check, and Zone B, baggage and body screening. Besides, we also estimate the average time needed in security check for those who belong to the trusted travelers of Pre-Check. The information is important for our simulations.

*1.2.    Our Work*

At the beginning, we design a simple model to illustrate the problems that ORD has in the busy seasons or days, by just assuming the flow of coming people to be approximately constant (small fluctuation is decided by a random number), and people tend to stand in the shortest line when they come. We find that the crowd are accumulating for the insufficient dealing ability of the check point, which means it is a saturation situation, and apparently that is why people always find long lines in the airports.

For the purpose of calculating the best solution for the crowded problem, we carefully design a statistic model, where statistic distribution is also considered for getting more precise outcomes. Moreover, more factors are also included in the model, such as the distribution of the check time, distribution of the coming-in flow, and so on. With the help of numerical simulations, we figure out the corresponding solution, after balancing plenty of factors that impact the average waiting time and waiting time variance.

We also refer to queueing theory to evaluate the efficiency of our model. Due to the complexity of model and pre-conditions, we decided not to calculate the analytical solutions to find the variance, but use numerical method instead to find the best solution for the airport.

More parameters are also included for describing different social customs, so that the model can be used in different countries or regions. We develop a new way to calculate the best operation for the airports, which positively enhanced the efficiency of security check, and minimize the cost or the investment of the airports. This method is suitable in many different airports around the world, because of the plenty parameters we have included. Evaluations are also provided in the last part of this paper. We point out the weakness and the strengths of the models we have created, and readers are welcome to offer their suggestions towards our models.

*1.3.    Assumptions*

- The flow of passengers is approximately constant when they come to the airport, and all of them are required to get through the security check.
- We only deal with the situations that the crowd are over the dealing ability of certain number of available check entrances, which means that no opening entrance is vacant during the busy days and busy hours.
- We assume that there are busy days and vacant days in the whole year. In busy days, more passengers and travelers are prepared to get aboard and leave for another places, such as summer vacation, the Christmas, winter vacation, etc. In these busy days, the airport is sure to face a more stressed situation and people also have to suffer from the long lines. So, we decide that the passenger current is constant in the whole day, but the daily total throughput varies from day to day because it may be busy or vacant.

2.    **Model**

*2.1.    Simple Model*

For ORD, the rate between regular security check entrances and Pre-Check entrances is 3: 1, and the amount of coming people is 2928 per hour, on average 0.81333 person per second. In this simple model we assume that each people would choose the shortest line when he starts waiting in the line. The total checking time, including document check in Zone A and screening in Zone B, is constantly set as 39 seconds, referring to the data from ORD. We use C++ to simulate the waiting process, and compute the average waiting time, and variance for everyone in the whole checking process.

Because of insufficient gates available for the waiting people, and on average each person needs 39 seconds to get through the security check, so a mass of congested crowd is inevitable in such an embarrassed situation, as shown in Fig.(1, 2, 3). As demonstrated in the figures, when the passengers are coming in constant flow, and only 4 gates are open for security check, which apparently indicates that there are not enough check entrances for such a busy passenger flow. For the simple reason that the dealing ability of all the available gates is weaker than the rate of new people coming into the lines, people cannot avoid the fact that get congested before the check point, wasting huge amount of time in vain. We think the current problems that most airports face can be described by this simple model.

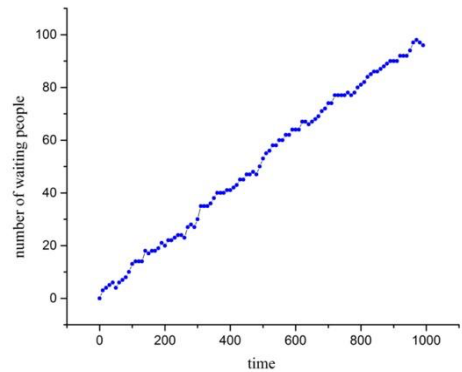

**Figure 1.** The number of waiting people that accumulates as time goes by, in the unit of second.

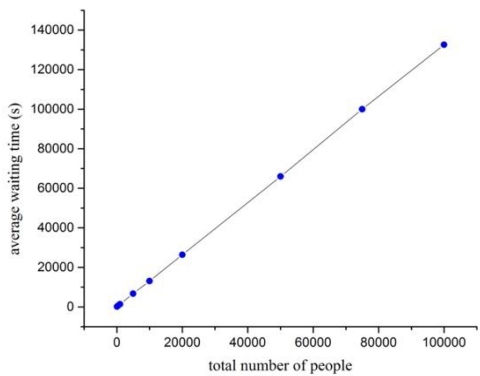

**Figure 2.** Average waiting time increases as total amount of people increases.

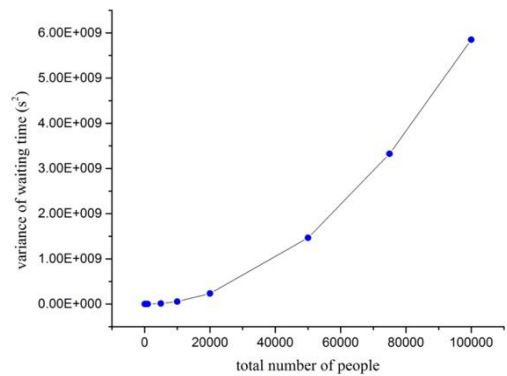

**Figure 3.** Variance of waiting time increases as total amount of people increases.

*2.2. Modifications*

We make a statistic model to describe the queueing problem and find the best solutions in this issue more precisely. The model mentioned above lacks many factors compared to the real situations, so that the former simple model cannot provide precise estimate for what is really happening in those crowded airports. Consequently, after referring to the queueing theory, we make a creative solution for solving the problem. We figure out how many gates are exactly needed when facing with different passenger flow. Hence, the staffs will easily know how many check entrances should be open when certain number of passengers are waiting for security check. In order to describe the true situation in the crowded airports, we add some factors:

- The statistic distribution of checking time for each person, include document check, baggage and body screening.
- The fluctuation of coming flow in a single day and in different days in a single year.
- The efficiency of security check.
- Number of serving check entrances.
- What influence of different queueing customs have to our model, such as various customs in America, China, Switzerland, or slower passengers.
- The cost needed for an airport.

## 2.3. Mathematical Theory

When passengers are coming for security check at a constant flow, say there are $\lambda$ passengers coming into one terminal each second, then the number of arrived passengers between $[0, t]$ follows Poisson distribution:

$$P\{X(t) = k\} = \frac{(\lambda t)^k e^{-\lambda t}}{k!}, \tag{1}$$

which means the possibility of finding $k$ people at time of $t$ is $P\{X(t) = k\}$.

Moreover, we carefully analyze what statistic rule controls the distribution of checking time. Considering time spent by screening devices are almost the same, and according to the distribution of screening time in ORD, we find that checking time follow Gaussian distribution:

$$P(a = x) = \frac{1}{\sqrt{2\pi}\sigma} e^{-\frac{(x-\mu)^2}{2\sigma^2}}, \tag{2}$$

where $\mu$ is the average security checking time for each passenger, and $\sigma^2$ is the variance of security check time for each passenger. For the case of ORD, we get $\mu = 39.02s$, $\sigma^2 = 191.68s^2$, then the standard deviation is $\sigma = 13.854s$. So obviously $1/\mu$ represents how many people can get through the security check for only one entrance, meanwhile $S/\mu$ represents how many people can get through when there are $S$ gates for security check.

In the simulation program, we use Monte Carlo 'acceptance rejection sampling' method to sample checking time, and in the algorithm checking time obeys Gaussian distribution.

Furthermore, by referring to queueing theory, when the number of passengers $n$ is larger than number of gates $S$, we can directly define

$$\rho = \frac{\lambda}{S}\mu \tag{3}$$

as the disposing ability of the security checkpoint. $\rho$ is the ratio between the increasing rate of amount needed to be disposed $\lambda$, and speed of disposing $S/\mu$. Or in another word, $\rho$ reflects the efficiency of the checking system. Apparently, with the balance state theory in the queueing theory, when $\rho < 1$, namely disposing ability $S/\mu$ is greater than $\lambda$, the so-called congested crowded line is not likely to appear.

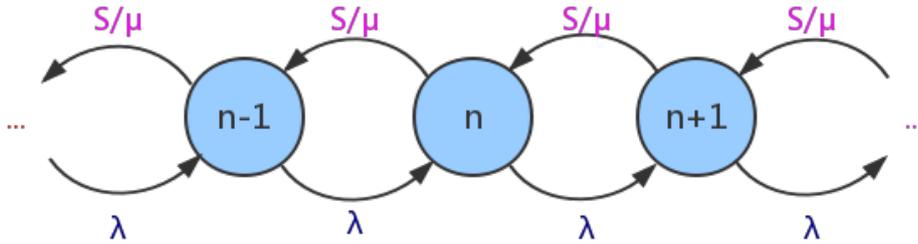

**Figure 4.** Diagram for state transitions.

On the other hand, regarding the state-transition equations in the queueing theory, the possibility of $n$ people waiting in lines is determined by equations

$$\begin{cases} (n+1)P_{n+1}\frac{1}{\mu} + \lambda P_{n-1} = \left(\lambda + \frac{n}{\mu}\right)P_n, & 1 \leq n \leq S; \\ \frac{S}{\mu}P_{n+1} + \lambda P_{n-1} = \left(\lambda + \frac{S}{\mu}\right)P_n, & n > S, \end{cases} \tag{4}$$

which means the state diverts as illustrated in Fig.(4).

Taking into the influence given by two types of lanes, the Pre-Check lane and regular lane, and in the saturation state, we define the incoming mission for Pre-Check lanes as $0.45\lambda$ and that for regular lanes as $0.55\lambda$. Let $S_p$ denote the number of opening Pre-Check gates, and $\mu_p$ the expectation of total check time for each person in Pre-Check lane. For regular check and Pre-Check, the gate number rate is $S/S_p = 3/1$, so the dealing time rate for each person is $\mu/\mu_p = 3 \cdot 45/55 = 2.454$.

### 2.4. Simulations

In this part we provide some result given by the numerical simulations, so that we can easily understand how to solve the problem most efficiently.

The coming flow is set as the normal situation, namely 2928 people coming per hour, so that on average 0.81333 person are coming per second. This passenger flow represents the average flow within the whole year. Most importantly, the algorithm uses Monte Carlo sampling to generate total checking time randomly, decided by the rules of Gaussian distribution. Still, people come and find the shortest line when they arrive.

We study the different situations when number of check gate varies from 12 to 28 in each terminal. The result of modeling is clearly shown in the figures. In the diagram we set the time unit in second (s) and variance in $s^2$.

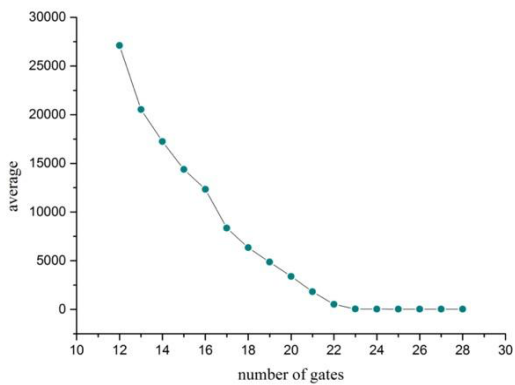
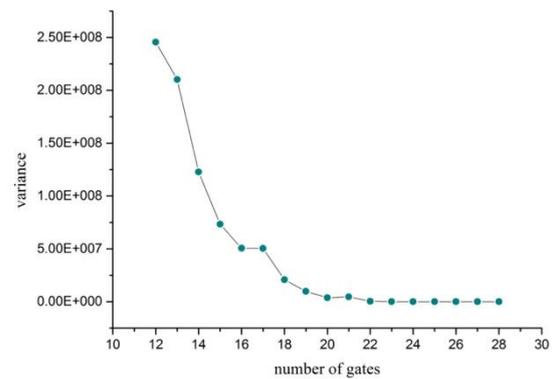

**Figure 5.** Relationship between average waiting time for every person, and number of gates, when passenger flow $\lambda$ is at average value 0.81333.

**Figure 6.** Relationship between variance of waiting time for every person, and number of gates, when passenger flow $\lambda$ is at average value 0.81333.

The balance solution for normal situation $\lambda = 0.81333$ is calculated in Section.2.3. The answer turns out to be that 24 gates would not cause congestion due to disposing ability is equal to flow value, as illustrated in Fig.(5,6). But is it wasteful to directly open 24 gates in each terminal?

The figures tell us the convincing fact that both average waiting time and variance decrease as more gates are available. In Fig.(6) the variance of waiting time for 20 to 28 gates is almost the same. In Fig.(5) average waiting time for 20 gates is about 3300 seconds, that is about 55 minutes, and average time for 21 gates is 1812 seconds, that is about 30 minutes. Hence, if the airport mainly wants to reduce variance in waiting lines, we reckon that 20 to 24 gates are just okay. Considering that people only need to wait half an hour when 21 gates are open, we therefore suggest that 21 is the optimal solution.

### 2.5. Optimal Solution

We also add financial cost into this model. Just assume that the cost of operating each gate, the equipment fee, the salary, electricity, maintaining fee are all the same for each gate and screening equipment, then we naturally draw the presumption that total cost of operating security checkpoint is linear to the total number of gates $S$, that is to say, the cost can be described as $C_{total} = C \cdot S$.

By using a common method of finding optimal solution in physics and engineering, if two variables have opposite tendency as certain parameter changes, we can calculate the product of two variables and look for the extremum value of the product. In this case, with more gates open for security check, the cost of operating the system becomes higher, but the average value of total waiting time $W_s$, and variance of total waiting time $V_s$ decrease. Therefore, we define the product of $C_{total}$ and $W_s$ as $Pro_A$ and the product of $C_{total}$ and $V_s$ as $Pro_V$:

$$Pro_A = C \cdot S \cdot W_s, \quad (5)$$
$$Pro_V = C \cdot S \cdot V_s. \quad (6)$$

Then we just need to find extremum value within reasonable range and figure out what the most appropriate gate number is when $\lambda = 0.81333$. In the computer program we simulate the process of waiting and draw the diagram for $Pro_A$ and $Pro_V$. Since $C$ has no influence on the solution, we can simply set $C = 1$.

Similar to Section.2.4, we simulate $Pro_A$ and $Pro_V$, as function of gate number. According to Fig.(5,6), we know that the best range is from 20 to 24 gates. After taking the cost into consideration, we find that the cost and fee do not have notable impact to the whole system, as $Pro_V$ is almost equal to each other when 20 to 24 gates are open, which reveals that 21 gates are suitable. Therefore, in the case of $\lambda = 0.81333$, after taking financial cost, average waiting time and waiting time variance into account, we finally draw the conclusion that 21 gates is the most appropriate solution, and the waiting time variance has been greatly decreased compared to former situation.

Likewise, we can also deal with other passenger flow in the same way. For instance, when it comes to larger passenger flow, like double the average flow, $\lambda = 1.62667$, after analyzing variance and waiting time, and $Pro_V$ as well, we decide that 44 gates are just okay. Fig.(7) shows $Pro_V$ when $\lambda$ comes to crest value.

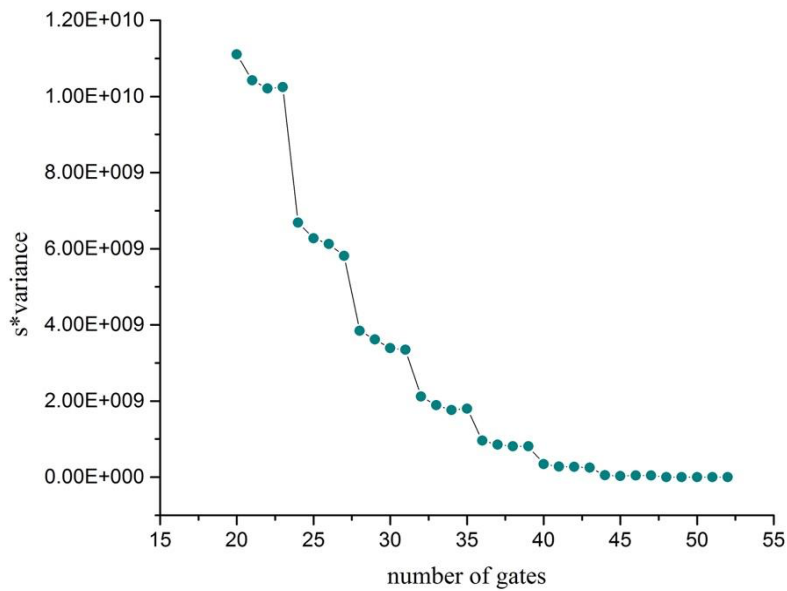

**Figure 7.** Relationship between $Pro_V$ and gate number $S$, when $\lambda = 1.62667$.

With the help of this model, in a similar way, we can also calculate the best solution when the flow value $\lambda$ is at other value. In our solution for ORD, the possibility of waiting two more hours is only 0.3%. It is really a good thing that we manage to reduce variance $V_s$ successfully, such that standard deviation of waiting time $\sigma_s$ is only 30 to 40 minutes, which guarantees 99.7% of passengers will never wait more than 2 hours. More than 70 percent of the passengers will finish security check at most one hour, and the average waiting time is only half an hour.

3. **Cultural Norms**

3.1. *Basic assumptions*

Different countries and regions have their own cultural norms that shape the local rules of social interaction. We introduce two more parameters to depict this issue: $P_c$, the percentage of people finding the shortest line as they arrive, and $E_l$, the error rate when people picking the shortest line. We pick the following parameter to simulate different cultures in different situations:

| Situations | $P_c$ | $E_l$ |
|---|---|---|
| Standard model | 1.00 | 0.00 |
| USA | 0.50 | 0.10 |
| China | 0.80 | 0.05 |
| Slower travelers | 0.00 | 0.05 |

The algorithm is shown in Fig.(8). All the judgement is made by random numbers in the algorithm.

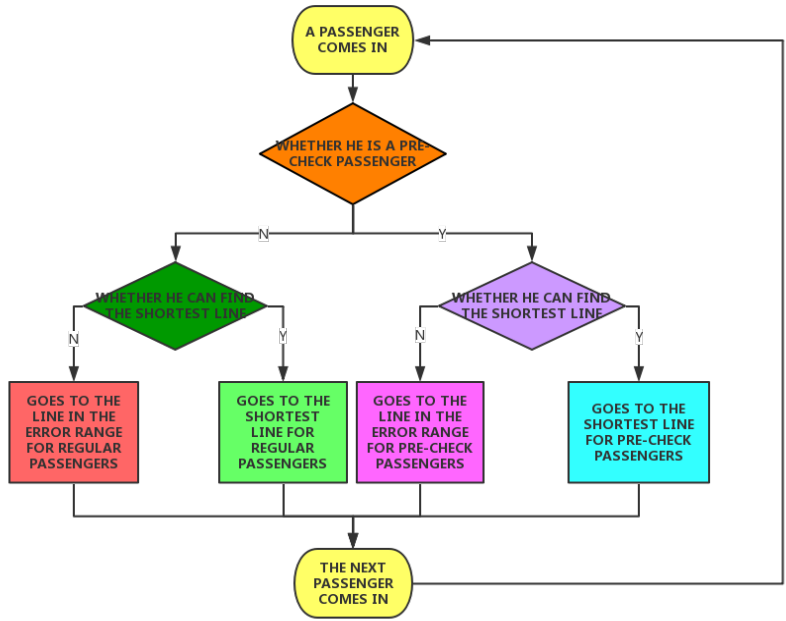

**Figure 8.** Process of the country-related model, also the main process of the algorithm.

3.2. *Simulations*

By utilizing the idea above, we study how different customs influence the amount of waiting people. Firstly, as shown in Fig.(9), which shows that except for slower travelers, the normal model, American and Chinese almost have same amount of waiting people when the input flow is fixed. We can easily draw the conclusion that if passengers are not interested in choosing the shortest time when they come in, the number of congested passengers increases greatly so that the efficiency gets lower compared to that of other models.

Besides, the slower traveler style can describe the place where the living pace is not fast, like North Europe, or South America. The model of America and China are suitable to describe fast-speed society, like London, Tokyo, Hong Kong or Singapore, etc.

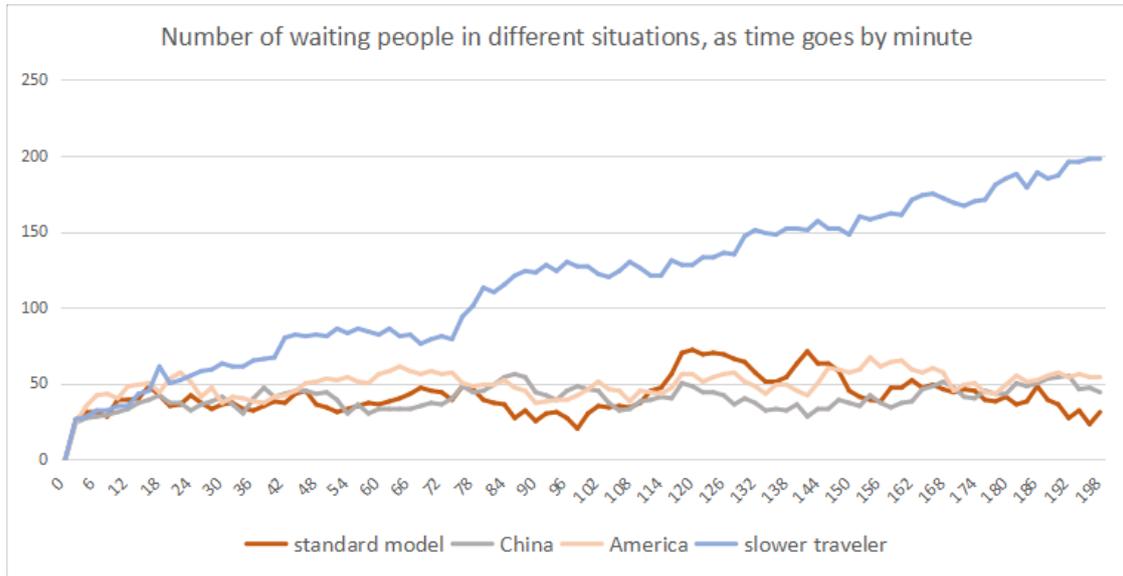

**Figure 9.** Number of waiting people, in different queueing style.

Furthermore, we are curious about the impact of these different customs and manners to the average waiting time $W_s$ and variance $V_s$. The custom's impact is shown in Fig.(10,11), for different gate number setup.

To our surprise, the different customs and manners have small effects on $W_s$ and $V_s$. Furthermore, we find that in Fig.(11), in the range of 20 to 23 gates, the variance is still the smallest at 21 gates. That is to say, the regional differences, like manners and norms, have little influence on choosing the number of gates $S$.

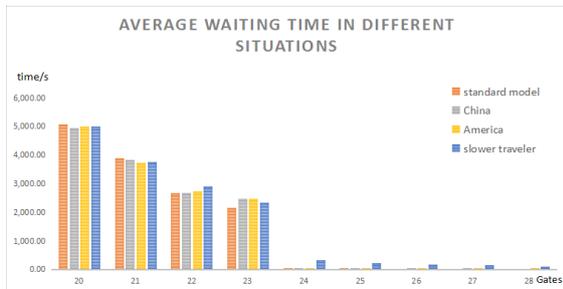

**Figure 10.** Average waiting time $W_s$ in different situations, as the gate number changes from 20 to 28.

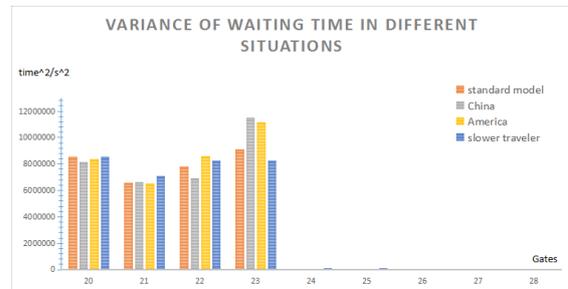

**Figure 11.** Variance of time $V_s$ in different situations, as the gate number changes.

## 4. Conclusions

We figure out the optimal solution to avoid endless waiting lines, by studying the data and security check setup for ORD. We mainly focus on the statistical rule the variables obey, so that we are able to find their influence more reasonably and precisely.

Our work points out the best solution when various passenger flows are coming and gives a relationship between the optimal entrance number and passenger flow, with the help of Monte Carlo method. Moreover, we also introduce the impact of different cultural norms to our model. By designing the parameters creatively, we exhibit how our model are affected by these additional factors. The solution we offer have greatly reduced the variance, for the standard deviation is only about 30 minutes, and the average waiting time is also half an hour. We prove that in our solution only 0.3% of the passengers may wait for more than 2 hours, even when the passenger flow is extremely large.